\documentclass[sigconf]{acmart}

\renewcommand\footnotetextcopyrightpermission[1]{} 
\AtBeginDocument{%
  \providecommand\BibTeX{{%
    \normalfont B\kern-0.5em{\scshape i\kern-0.25em b}\kern-0.8em\TeX}}}

\usepackage{multirow}
\usepackage{textcomp} 
\usepackage{ulem}

\begin{document}

\title{AI and the Sense of Self}

\author{Srinath Srinivasa}
\email{sri@iiitb.ac.in}
\orcid{0000-0001-9588-6550}
\author{Jayati Deshmukh}
\email{jayati.deshmkh@iiitb.org}
\affiliation{%
  \institution{International Institute of Information Technology}
  \streetaddress{26/C, Electronics City Phase 1}
  \city{Bangalore}
  \state{Karnataka}
  \country{India}
  \postcode{560100}
}

\renewcommand{\shortauthors}{Srinivasa, et al.}

\begin{abstract}
After several winters, AI is center-stage once again, with current advances enabling a vast array of AI applications. This renewed wave of AI has brought back to the fore several questions from the past, about philosophical foundations of intelligence and commonsense-- predominantly motivated by ethical concerns of AI decision-making. In this paper, we address some of the arguments that led to research interest in intelligent agents, and argue for their relevance even in today's context. Specifically we focus on the cognitive sense of ``self'' and its role in autonomous decision-making leading to responsible behaviour. The authors hope to make a case for greater research interest in building richer computational models of AI agents with a sense of self. 
\end{abstract}

\ccsdesc[500]{Computing methodologies~Philosophical/theoretical foundations of artificial intelligence}

\keywords{machine ethics, autonomous agents, reinforcement learning, sense of self}

\maketitle
\pagestyle{plain}
\section{Introduction}
``Artificial Intelligence'' or AI, a term coined by Minsky and McCarthy in 1956\footnote{History of Artificial Intelligence\\ https://en.wikipedia.org/wiki/History\_of\_artificial\_intelligence}, has evolved into a veritable global vision and dream, evoking interest from not just researchers, but also practitioners, artists, writers, policy makers, and the general public. Like most fields of study, AI has gone through several waves interspersed by periods of relative insignificance-- the so called ``AI winters.'' 

Each wave of AI resurgence has been characterized by specific forms of conceptual advancements-- like formal logic, artificial neural networks, intelligent agents, subsumption architecture, etc. The current resurgence in interest in AI is perhaps unique in that regard, since arguably, the primary catalyst for this new wave comes from advances in hardware, especially Graphical Processing Units (GPUs), re-purposed for massive parallel processing of Artificial Neural Networks. This wave is hence driven more by AI applications and deployments, rather than by conceptual breakthroughs. Although in the last decade, there have been several new advances in deep-learning architectures, autonomous agents and robotic interaction models, arguably, none of them constitute a paradigmatic departure from earlier models. 


This also implies that much of the open questions and challenges posed by AI from earlier times, have remained unanswered. Of significance is the issue of \textit{machine ethics}, which was once primarily a philosophical debate, but which has now become center-stage with large-scale deployment of AI in different application contexts. 

Machine ethics refers to  a family of disparate concerns. With data-heavy applications like recommendation and personalization systems, biases in the data or in the algorithmic design assumptions, can pose several ethical concerns~\cite{kirkpatrick2016battling,hajian2016algorithmic,vorvoreanu2019gender}. Similarly, AI agents acting autonomously in order to achieve some objective, may create several kinds of collateral damage with serious ethical implications~\cite{goodall2014machine,tolmeijer2020implementations}. 

For a large part, ethical considerations for machines have been specified using normative constructs, and modeled as either a constraint satisfaction problem or a constrained optimization problem. Different paradigms are used to model underlying ethical guidelines. These include~\cite{tolmeijer2020implementations}: deontics (specification of what one ought to do), consequentialism (reasoning based on expected consequences), modeling of an innate sense virtues in agents, and particularism (context-specific ethical reasoning). In addition, emerging areas like Artificial Moral Agents (AMA), Reflective Equilibrium (RE), and Value Sensitive Design (VSD) have addressed formal modeling of ethical frameworks as a fundamental design principle of AI systems~\cite{jobin2019global,daniels1979wide,floridi2004morality,friedman1996value,friedman2002value,fossa2018artificial,wallach2008machine,wallach2010conceptual}. 

However, this paper argues that our understanding of machine ethics is far from complete, and that there is a need to reopen some of the philosophical debates from the 1980s and 1990s about the nature of intelligence, and address them in today's context. There are fundamental issues with the way ``intelligence'' is defined and modeled in present day AI systems, that create a barrier for AI to reason about ethics seamlessly. Ethics and intelligence are often assumed to be orthogonal, if not conflicting dimensions. 


Most questions pertaining to modeling ethics, require some form of generalized understanding of ethical principles, necessitating an element of ``commonsense'' reasoning~\cite{powers2006prospects}-- leading to yet another long-standing open issue, that is typically relegated to ``strong'' AI or Artificial General Intelligence (AGI). 

Many such considerations led to the emergence of the field of Intelligent Agents (IA), addressing issues like agency, autonomy, self-interest, and so on. Multi-Agent Systems (MAS) extended on this concept, to model interacting autonomous agents and their emergent properties. The field of multi-agent systems have had to contend with issues of ethics and responsibility, when self-interest from disparate agents interfere with one another. This lead to the development of several forms of multi-agent negotiation protocols and fairness constructs~\cite{kraus1997negotiation,panait2005cooperative,vidal2003multiagent}.

While IA and MAS elicited a lot of research interest in the early 2000s, the interest soon waned. This paper tries to bring back some of the key philosophical arguments that lead to research interest in computational modeling of agency, with the hope that some of them may provide promising paths of inquiry for some of the pressing concerns of AI deployments today. 

In particular, the authors propose to extend on some of the arguments around agency, and propose that an ``elastic sense of self'' is a key ingredient that can address disparate issues concerning self-interest, ethics and responsible behaviour. 

\section{Machines and Societies}\label{sec:machinessocieties} 

Scientific and engineering models today are predominantly grounded in Newtonian hermeneutics, where reality is considered to be built from impersonal, inanimate matter, and causal relationships between them. This form of thinking replaced earlier models of human inquiry, that were overtly anthropomorphic. Hence for instance, we no longer consider an earthquake today as an expression of ``anger'' of some God, but as a causal chain of tectonic events leading to the catastrophe. 

Newtonian hermeneutics has enabled us to build rich causal models of physical phenomena, paving the way for machines and robots that are as versatile as natural beings, if not more, as regards their mechanical abilities. However, when such machinery needs to inter-operate in an ecosystem of sentient beings like humans and animals, they pose great challenges, since there is no place for anthropomorphic constructs like free-will, conscience, trust, desire, anger, etc. as part of Newtonian hermeneutics. This makes a lot of social constructs and communication paradigms inaccessible and inapplicable to machines. For instance, ``shaming'' or expressing disapproval, disgust and anger against a reprehensible act can act as a deterrent to a human; but machines hitherto don't respond to such expressive rhetoric. While we can appeal to the \textit{conscience} of a human wrong-doer to make them correct their actions, no such mechanisms exist for interacting with AI that is about to do something irresponsible. 

To some extent, present day AI can be made to respond to rhetoric, by modeling them as reinforcement signals from the environment. Indeed, in a number of AI deployments, ethical issues are enforced by means of constraints and/or reinforcements over an underlying adaptive logic~\cite{abel2016reinforcement,noothigattu2019teaching}. 

But this only opens up deeper questions about how to bring about ethical and responsible behaviour in the absence of relevant reinforcements and constraints. A sense of ethics in humans are not always responses to external reinforcements. Appeal to conscience of a person, is not the same as deterrence by inducing the fear of penalty. Indeed, the system of external reinforcements in the form of laws and social norms, are themselves an emergent characteristic of complex interactions around ethics, among humans. There is evidence that a sense of ethics and responsible behaviour is an \textit{innate} element of human nature~\cite{bregman2020humankind,hamlin2013moral}.

This leads us to ask whether there are some paradigmatic differences between natural beings and artificial automation. Could it be that the idea of machine ethics is itself an ill-posed problem? Could it be that machines today, lack essential design elements that are present in natural beings, and that which endow them with anthropomorphic abilities including their sense of ethics? 

When we compare automation in nature, and that of human engineering, we can immediately see a number of contrasts between an artificial machine (like a car), and a natural being (like a tiger)-- even in their mechanics. 

Firstly, we can see that \textit{nature does not have wheels}, and hardly if ever, bases its mechanics on rotary motion. There are hardly any examples of motors, pumps and turbines in nature that are based on rotary motion. In contrast, the wheel is such a fundamental element of human engineering, that ``don't reinvent the wheel'' is an oft-repeated clich\'{e}. But, nature has not ``invented'' the wheel at all! 

Natural pumps, like the heart in animals, use contraction and expansion as a means for pumping. Conventional mechanical engineering would call that an inefficient design, since continuous contraction and expansion leads to material fatigue and wear-and-tear. However, the heart beats continuously throughout the lifetime of the natural being (60-100 years for average humans), without ever taking a break-- a feat that is very hard if not impossible to achieve with more ``efficient'' motors built from conventional engineering! 

Clearly, paradigmatic differences between natural and artificial engineering can lead to fundamental differences in engineering wisdom. What is clearly an unwise design in one paradigm, is the design of choice in the other. Could this paradigmatic difference hold the key for us to understand anthropomorphic constructs that are an integral part of natural beings, but not of artificial machines? 

The paradigmatic difference between artificial and natural automation can be summarized as the difference between a ``machine'' and a ``society.''

Unlike machines, which are built from components custom-made for their functionality, natural beings are modeled as a large ``society'' of autonomous entities called cells. Cells are generic components, which in their nascent stages (called stem cells), can be moulded into several different kinds of functional agents, like muscles, nerves, cartilage, bone matter, tissue, etc.

The master-plan directing role distribution among the cells is encoded in an organism's genotype, or its genetic material. But unlike the ``blueprint'' of machines, a genotype does not rigidly encode the phenotype (the resultant organism). The structure and function of the phenotype is a combination of both genetically encoded plan, and adjustments to its environment (nature and nurture). 

The logic that drives moulding of stem cells into specific forms of functional roles is based on the \textit{economic demand} from the ``society'' that makes up the organism. Hence, a physically active organism creates a larger demand for muscle cells to develop, much like growth in a particular sector (like say, biotechnology) of a human society, creates a demand for more professionals to be trained in this area. 

Machine-oriented and society-oriented designs lead to some sharp differences in engineering wisdom, as noted earlier. Society-oriented design needs to work with building blocks that are autonomous, and act independently in their individual interest. For the system of cells to work together as an organism, the system needs to be such that acting in cooperation with other cells is far more rationally lucrative than acting independently.

Cooperation does not mean that the collective will always overrides individual autonomy. The autonomous nature of individual components makes organisms immensely adaptive and self-sustaining. Organisms have intelligent responses encoded throughout their being. This \textit{pervasive intelligence} of societies result in resilience and self-healing properties like responding to routine issues like a scratch or a skin prick, in a subconscious manner-- without sometimes the brain (representing the collective society) even being aware of it. 

But by far, a characteristic feature of natural beings that has been largely ignored by engineers, is the sense of ``self'' that pervades across all cells of the organism. Cells have a sharp notion of ``citizenship'' to the being that make them act with vigilance against ``foreign'' cells that infect the organism. Even though each agent in the being is acting autonomously, there is also a sense of ``oneness'' or ``belongingness'' to the being, that pervades across all the agents. 

When the organism strives to survive, it is the pervasive sense of self that is sought to be maintained and preserved and not for instance, any particular cell. It is also the pervasive sense of self, that is sought to be protected against attacks in the form of infections, by the immune system. 

The sense of self is also ``elastic'' in the sense that the being may sometimes \textit{identify} with other external entities or concepts, by attaching a part of its sense of self, to that object. Identifying with an external entity means that, the being contributes some part of its biological and cognitive processes towards preserving and furthering the interests of the object of identity. 

Hence, parents identifying with their children, or patriots identifying with their country, or activists identifying with a cause, proactively invest their efforts and mind towards the interests of their object of identity. This elastic sense of self may also underlie the mirror neuron system (MNS) that is thought to be the neurological basis for empathy~\cite{rizzolatti2005mirror,oberman2007human}. 

We argue that modeling this \textit{elastic sense of self}, holds the key for several issues pertaining to responsible AI, and hope to elicit more research interest in this area. 

\section{Computational Modeling of  Agency}\label{sec:autonomy} 

Computational modeling of agency and autonomy began to elicit increasing research interest starting from the late 1980s. A survey of different computational models of agency may be found in~\cite{srinivasa2020evolution}. 

Early models of agency focused on the proactive nature of autonomous agents, implemented as software objects with an independent thread of execution. Later on, logics based on intentionality and norms, tempered by an agent's beliefs and knowledge, were developed for modeling autonomy~\cite{georgeff1998belief,rao1991modeling}. A third paradigm of agency were adaptive models powered by reinforcement learning and extensive games~\cite{lin1992self,shoham1993agent}. 

While the above approaches resulted in rich, proactive and adaptive behaviour, questions still remained about what is meant by autonomy itself. Perhaps the closest we have come to answering this question is to model autonomy using the theory of \textit{rational choice}~\cite{panait2005cooperative,boella2002game}. Rational choice is represented by two elements, \textit{self-interest}, and \textit{utility maximization}. 

Foundations of rational choice and economic games come from the work of von Neumann and Morgenstern~\cite{morgenstern1953theory}, that is also now called the ``classical'' model of rational choice. This theory is based on representing self-interest in the form of preference functions between pairs of choices. Ordinal preference relations between pairs of choices, are converted to numerical payoffs based on equating expected payoffs of a conflicting set of choices. For instance, suppose an agent prefers $A$ over $B$ over $C$ (represented as $A > B > C$). Suppose now that the agent is presented a choice, where Choice I returns $B$ with 100\% certainty, and Choice II returns either $A$ or $C$ with a probability of $p$ and $(1-p)$ respectively. The value of $p$ at which the agent becomes indifferent between choices I and II provides us a mechanism for assigning numerical payoff values to $A$, $B$ and $C$. The classical theory is also developed further, with a set of axiomatic basis like methodological individualism, transitivity of preferences, independence of choices, etc. 

While classical rational choice theory is widely used in different application areas, including modeling human behaviour and micro-economic models, it has also received criticism from various quarters about how well it can model human sense of agency. In his critique called ``Rational Fools''~\cite{sen1977rational}, Sen argues that our autonomy comes from our ``sense of self'' and it is too simplistic to reduce our sense of self to a preference matrix between pairs of choices. Specifically, Sen argues that humans display an innate sense of trust and empathy towards others, and assume a basic level of trust to exist even among self-interested strangers. If humans were to be strict rational maximizers, then according to Sen, the following kinds of interactions would be more commonplace~\cite{sen1977rational}: 
\begin{quote}
    "Where is the railway station?" he asks me. "There," I say, pointing at the post office, "and would you please post this letter for me on the way?" "Yes," he says, determined to open the envelope and check whether it contains something valuable.
\end{quote}

This critique of the classical model lead to the development of the theory of rational empathy and welfare economics. 

Similarly, Kahnemann and Tversky in their work called ``prospect theory''~\cite{kahneman2013prospect}, critique the classical model for its linear model of utility from expected payoffs. 

\begin{figure}
    \centering
    \includegraphics[width=3.5in]{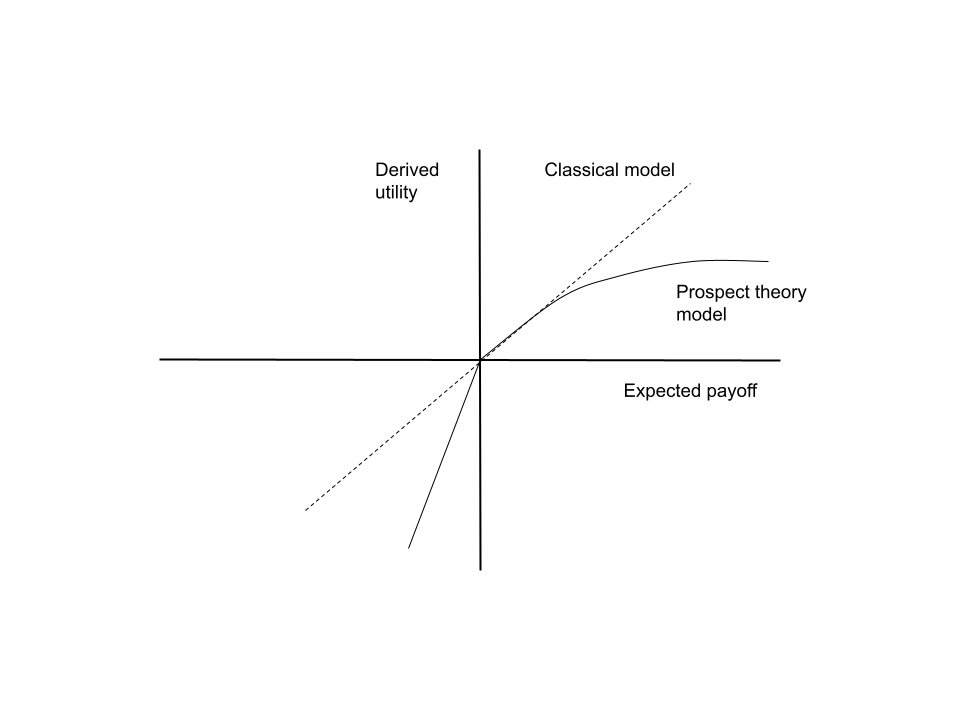}
    \caption{Contrasting derived utility between classical model and prospect theory}
    \label{fig:prospect_th}
\end{figure}

Figure~\ref{fig:prospect_th} contrasts the model of derived utility between the classical model and prospect theory. Utility is also called ``intrinsic payoff'' and refers to the value associated by the agent to an external payoff received. 

Prospect theory identifies at least two characteristics of human valuation to external payoffs-- \textit{saturation}, and \textit{risk aversion}. Saturation refers to diminishing valuation of returns with increasing returns. The first million earned may be valued very highly, by a business, but by the time the business earns 50 million, it is largely business as usual. 

Similarly, humans value negative and positive payoffs differently. Humans are known to be ``risk averse'' and value prospects of negative returns much more negatively, than positive returns of the same worth. Hence an investment that provides a guaranteed return of $x$ is valued higher than another investment that returns either $0$ or $2x$ with equal probability. 

Both saturation and risk-aversion can also be tagged back to our ``sense of self''. Risk aversion comes from our pursuit of \textit{homeostasis}-- preserving our sense of self, making us more often to choose smaller but guaranteed returns, over higher but riskier returns. Similarly, saturation can be explained by our mind's eternal quest for novelty or epistemic surprise~\cite{clark2018nice}, where unexpected rewards are valued more than expected returns. 

\section{An Elastic Sense of Identity}\label{sec:identity} 

It is reasonably clear that the human sense of autonomy is much more than rational choice, as described by the classical model. Critiques of the classical model introduce several facets of our sense of self, including: rational empathy, trust, homeostasis, foraging or epistemic novelty, risk aversion, etc. 

While we may be far from a comprehensive computational model of self, in this work, we focus on a specific characteristic of our sense of self that may hold the key for the innate sense of responsibility and ethics in humans. We call this the \textit{elastic} sense of self, extending over a set of external objects called the \textit{identity set}. 

Our sense of self, is not limited to the boundaries of our physical being, and often extends to include other objects and concepts from our environment. This forms the basis for social identity~\cite{jenkins2014social} that builds a sense of belongingness and loyalty towards something other than, or beyond one's physical being. 

We model this formally as follows. Given an agent $a$, the sense of self of $a$ is described as: 
\begin{equation}
    S(a) = (I, d_a, \gamma_a)
\end{equation}

Here $I$ is the set of \textit{identity objects}, where $a \in I$. The agent itself belongs to its set of identity objects. The set may contain any number of other entities including other agents, collections of agents, or even abstract concepts. The term $d_a: \{a\} \times I \rightarrow \Re^+$ represents the ``semantic distance'' between $a$ and some object in its identity set, with $d_a(a) = 0$. The term $\gamma_a \in [0,1]$ represents an attenuation parameter, indicating how fast does the sense of identity attenuate with distance. The agent identifies with object at distance $d$ with an attenuation of $\gamma_a^{d}$. 

The ``sense of self'' of the agent describes how its internal valuation or \textit{utility}, is computed based on external rewards or \textit{payoffs} that may be received by elements of its identity set. For any element $o \in I$, let the term $o_i$ refer to the payoff obtained by object $o$ in game (system) state $i$. 

Given this, the utility derived by agent $a$ is computed as follows: 
\begin{eqnarray}
    u_i(a) & = & \frac{1}{Z} \displaystyle \sum_{\forall o \in I} \gamma_a^{d_a(o)} o_i
    \label{eqn:utility1} \\
    Z & = & \displaystyle \sum_{\forall o \in I} \gamma_a^{d_a(o)}
    \label{eqn:utility2}
\end{eqnarray}

The above can be understood as a ``unit'' of self being attached in different proportions to the objects in the identity set, based on their semantic distance and attenuation rate. Since the distance from an agent to itself is zero, this has the least attenuation. 


\begin{table}[th]
    \centering
	\begin{tabular}{cc|cc}
	& & \multicolumn{2}{c}{Player A} \\
	& & C & D \\
	\hline
	\multirow{2}{*}{Player B} & C & 6, 6 & 0, 10 \\
	& D & 10, 0 & 1, 1 \\
	\end{tabular}
    \caption{Prisoner's Dilemma}
    \label{tab:pd}
\end{table}

To illustrate the impact of an elastic sense of identity, consider the game of Prisoners' Dilemma as shown in Table~\ref{tab:pd}. 

The Prisoners' Dilemma (PD) represents a situation where players have to choose to cooperate (C) or defect (D) on the other. When both players cooperate, there are rewarded with a payoff (6 in the example). However, as long as one of the players chooses to cooperate, the other player has a temptation to defect, and end up with a much higher payoff (10 in the example). Hence, a player choosing to cooperate, runs the risk of getting exploited by the other player. And when both players choose to defect on the other, they end up in a state of ``anarchy'' with a much lesser payoff (1 in the example), than had they both chosen to cooperate. 

When played as a one-shot transaction, there is no rational incentive for a player to choose to cooperate. Regardless of whether a player is known to choose cooperate or defect, it makes rational sense for the other player to choose $D$ over $C$. The state $DD$ is also the Nash equilibrium, representing the mutual best response by both players, given the choice of the other. The choice $D$ strictly dominates over choice $C$, since regardless of what the other player chooses, a player is better off choosing $D$ over $C$. 

The only way players in a PD game find a rational incentive to cooperate, is when the game is played in an \textit{iterated} manner, with evolutionary adjustments allowing players to change strategies over time~\cite{axelrod1981evolution}. 

However, with an elastic sense of identity, we can create a rational incentive for the players to cooperate, even in a one-shot transaction. Instead of working on strategies and payoffs, here we change the players' \textit{sense of self}, to include the other player, to different extents. 

Without loss of generality, consider player $A$, and let the payoff in game state $i$ be denoted as $a_i$. With an elastic identity that includes the other player in one's identity set at a distance of 1, the derived utility of player $A$ in game state $i$ is given by (from Eqn~\ref{eqn:utility1}): 
\begin{equation}
    u_i(A) = \frac{1}{1+\gamma} [a_i + \gamma b_i]
\end{equation}

\begin{figure}
    \centering
    \includegraphics[width=3.5in]{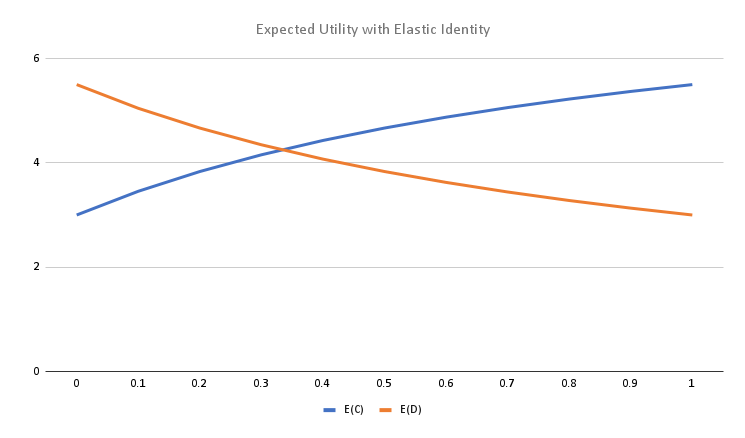}
    \caption{Change in expected utility with increased elasticity of sense of self}
    \label{fig:elasticidenti}
\end{figure}

The \textit{expected utility} of a choice (either $C$ or $D$) is computed by the utility accrued at all possible game states on making this choice, along with the probability of this game state. Since we make no further assumptions, all game states are considered equally probable. Hence, the expected utility for choosing a given choice is computed as follows: 

\begin{eqnarray}
    E_A(C) = 0.5 \cdot u_{CC}(A) + 0.5 \cdot u_{CD}(A) \\
    E_A(D) = 0.5 \cdot u_{DC}(A) + 0.5 \cdot u_{DD}(A)
\end{eqnarray}

Figure~\ref{fig:elasticidenti} plots the expected utility from choosing $C$ or $D$ over varying values of $\gamma$ or the elasticity in one's sense of self. When $\gamma = 0$, this becomes the usual PD game, where the expected utility from choosing $D$ is much higher than the expected utility from choosing $C$. However, as $\gamma$ increases, with player $A$ identifying more and more with player $B$, the expected utility of choosing $C$ overtakes that of choosing $D$, when $\gamma = \frac{1}{3}$. At this value of $\gamma$ the sense of self is split between one's own interest and the other's interest, in a ratio of $\frac{3}{4} : \frac{1}{4}$.

When $\gamma = 1$ the sense of self is evenly split between a player and the other. In this state, the PD game effectively ``flips over'' with $C$ and $D$ swapping places with respect to expected utility. In this state, it makes as much rational sense to choose to cooperate, as it made rational sense in the conventional PD to choose to exploit the other. This seems to give credence to the folk wisdom that a relationship between two persons is at its natural ideal when each member feels as much for the other, as for themselves. In such a state, cooperation is more appealing than selfish gains.

An elastic sense of self can be contrasted with other forms of pro-social constructs that have inspired the design of fairness in artificial agents. We look at a few of these constructs here. 

\paragraph{Pareto Optimality} One of the  commonly used constructs for fairness is Pareto optimality~\cite{banerjee2007reaching,de2008artificial}. A game state is said to be Pareto optimal, if it does not contain any ``Pareto improvement'' where an agent can improve one's payoffs by changing its choice, such that it does not result in a reduced payoff for any other agent. 

\begin{figure}
    \centering
    \includegraphics[width=2.75in]{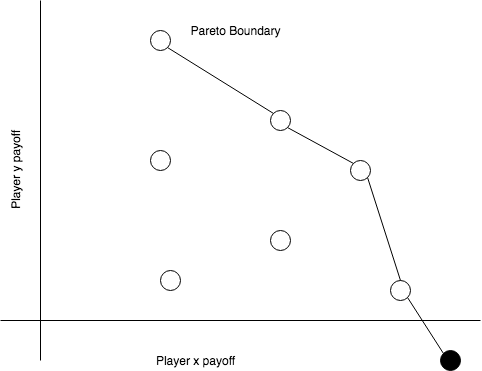}
    \caption{Pareto boundary and fairness}
    \label{fig:pareto}
\end{figure}

In the PD example, while game state $CC$ is not a Nash equilibrium, it is Pareto optimal, since no player can switch to the other choice to get a better payoff, without hurting the other. In this sense, ``consideration for the other'' or ``aversion to inequity in incremental payoffs'' as an ethical principle, can lead to cooperation. 

However, Pareto optimality can also just as well result in grossly unfair configurations. Figure~\ref{fig:pareto} shows a two player game with several states. The set of states in the ``Pareto boundary'' connected by the line, represent Pareto optimal states. In these states, no player can change its choice to get a better payoff for itself, without hurting the other. As we can see, the shaded state, where Player $y$ has a negative payoff, is also on the Pareto boundary! 

Pareto optimality, used by itself as a measure of responsible behaviour, can also admit oppressive constructs as ``fair'' configurations.

\paragraph{Altruism} Altruism or ``selfless'' behaviour where an agent ``sacrifices'' one's own good for the welfare of the other, or for the collective, is sometimes celebrated as the epitome of pro-social responsibility. 

However, as we can see from the PD game, while an altruist would be attracted by the state $CC$, the game state $CD$ would also be attractive for an altruist agent, since this gives the best possible payoff for the other player. 

Not being concerned about one's own welfare, doesn't necessarily lead to responsible behaviour. Consider a surgeon or pilot sacrificing their sleep or rest time to maximally serve their patients or passengers. They would be putting them at risk rather than serving them. \\

An elastic sense of self on the other hand, does not put one's own self in conflict with the interests of others, nor does it invalidate one's own individuality for collective interests. 



\section{Conclusions} 

The objective of this paper is to address the question of machine ethics in a philosophical manner using foundations of human cognition, and propose a new line of thinking for modeling responsibility in AI agents. An elastic sense of self, as proposed in this paper, may be a foundational element for modeling several forms of anthropomorphic constructs in AI, including machine ethics. 

For future work, we plan to take up realistic agent-based applications involving reinforcement learning, and introduce an elastic sense of self into the learning agents. Preliminary work in this regard has yielded promising results, with agents pursuing self-interest while being mindful of collateral damage. 

Elastic identity also opens up several new questions and opportunities for research. Some of these questions include the following: underlying model for deciding the elements of one's identity set, semantic distance to each element of one's identity set, and the attenuation parameter. There are also questions about how and when is the attenuation parameter set, and whether it changes over time and with fruitful or adverse experiences. 

At a systemic level, there are also open questions about the evolutionary stability of a system of agents with elastic identity. Can a system of empathetic agents be successfully ``invaded'' by a small group of non-empathetic agents who don't identify with others? Or does there exist a strategy for deciding the optimal level of one's empathy or extent of one's identity set, that makes it evolutionarily stable?

\bibliographystyle{ACM-Reference-Format}
\bibliography{ref}

\end{document}